# Supervised Transformer Network for Efficient Face Detection

Dong Chen, Gang Hua, Fang Wen, and Jian Sun

Microsoft Research
{doch,ganghua,fangwen,jiansun}@microsoft.com

**Abstract.** Large pose variations remain to be a challenge that confronts real-word face detection. We propose a new cascaded Convolutional Neural Network, dubbed the name Supervised Transformer Network, to address this challenge. The first stage is a multi-task Region Proposal Network (RPN), which simultaneously predicts candidate face regions along with associated facial landmarks. The candidate regions are then warped by mapping the detected facial landmarks to their canonical positions to better normalize the face patterns. The second stage, which is a RCNN, then verifies if the warped candidate regions are valid faces or not. We conduct end-to-end learning of the cascaded network, including optimizing the canonical positions of the facial landmarks. This supervised learning of the transformations automatically selects the best scale to differentiate face/non-face patterns. By combining feature maps from both stages of the network, we achieve state-of-the-art detection accuracies on several public benchmarks. For real-time performance, we run the cascaded network only on regions of interests produced from a boosting cascade face detector. Our detector runs at 30 FPS on a single CPU core for a VGA-resolution image.

## 1 Introduction

Among the various factors that confront real-world face detection, large pose variations remain to be a big challenge. For example, the seminal Viola-Jones [1] detector works well for near-frontal faces, but become much less effective for faces in poses that are far from frontal views, due to the weakness of the Haar features on non-frontal faces.

There were abundant works attempted to tackle with large pose variations under the regime of the boosting cascade advocated by Viola and Jones [1]. Most of them adopt a divide-and-conquer strategy to build a multi-view face detector. Some works [2–4] proposed to train a detector cascade for each view and combine their results of all detectors at the test time. Some other works [5–7] proposed to first estimate the face pose and then run the cascade of the corresponding face pose to verify the detection. The complexity of the former approach increases with the number of pose categories, while the accuracy of the latter is prone to the mistakes of pose estimation.

Part-based model offers an alternative solution [8–10]. These detectors are flexible and robust to both pose variation and partial occlusion, since they can reliably detect the faces based on some confident part detections. However, these methods always require the target face to be large and clear, which is essential to reliably model the parts.

Other works approach to this issue by using more sophisticated invariant features other than Haar wavelets, *e.g.*, HOG [8], SIFT [9], multiple channel features [11], and high-level CNN features [12]. Besides these model-based methods, Shen *et al.* [13] proposed to use an exemplar-based method to detect faces by image retrieval, which achieved state-of-the-art detection accuracy.

It has been shown in recent years that a face detector trained end-to-end using DNN can significantly outperforms previous methods [10, 14]. However, to effectively handle the different variations, especially pose variations, it often requires a DNN with lots of parameters, inducing high computational cost. To address the conflicting challenge, Li *et al.* [15] proposed a cascade DNN architecture at multiple resolutions. It quickly rejects the background regions in the low resolution stages, and carefully evaluates the challenging candidates in the high resolution stage.

However, the set of DNNs in Li *et al.* [15] are trained sequentially, instead of end-to-end, which may not be desirable. In contrast, we propose a new cascade Convolutional Neural Network that is trained end-to-end. The first stage is a *multi-task* Region Proposal Network (RPN), which simultaneously proposes candidate face regions along with associated facial landmarks. Inspired by Chen *et al.* [16], we jointly conduct face detection and face alignment, since face alignment is helpful to distinguish faces/non-faces patterns.

Different from Li *et al.* [15], this network is calculated on the original resolution to better leverage more discriminative information. The alignment step warps each candidate face region to a canonical pose, which maps the facial landmarks into a set of canonical positions. The aligned candidate face region is then fed into the second-stage network, a RCNN [17], for further verification. Note we only keep the $K$ face candidate regions with top responses in a local neighborhood from the RPN. In other words, those Non-top K regions are suppressed. This helps increase detection recall.

Inspired by previous work [18], which revealed that joint features from different spatial resolutions or scales will improve accuracy. We concatenate the feature maps from the two cascaded networks together to form an architecture that is trained end-to-end, as shown in Figure 1. Note in the learning process, we treat the set of canonical positions also as parameters, which are learnt in the end-to-end learning process.

Note that the canonical positions of the facial landmarks in the aligned face image and the predicted facial landmarks in the candidate face region jointly defines the transform from the candidate face region. In the end-to-end training, the training of the first-stage RPN to predict facial landmarks is also supervised by annotated facial landmarks in each true face regions. We hence call our network a Supervised Transformer Network. These two characteristics differentiate our model from the Spatial Transformer Network [19] because a) the Spatial

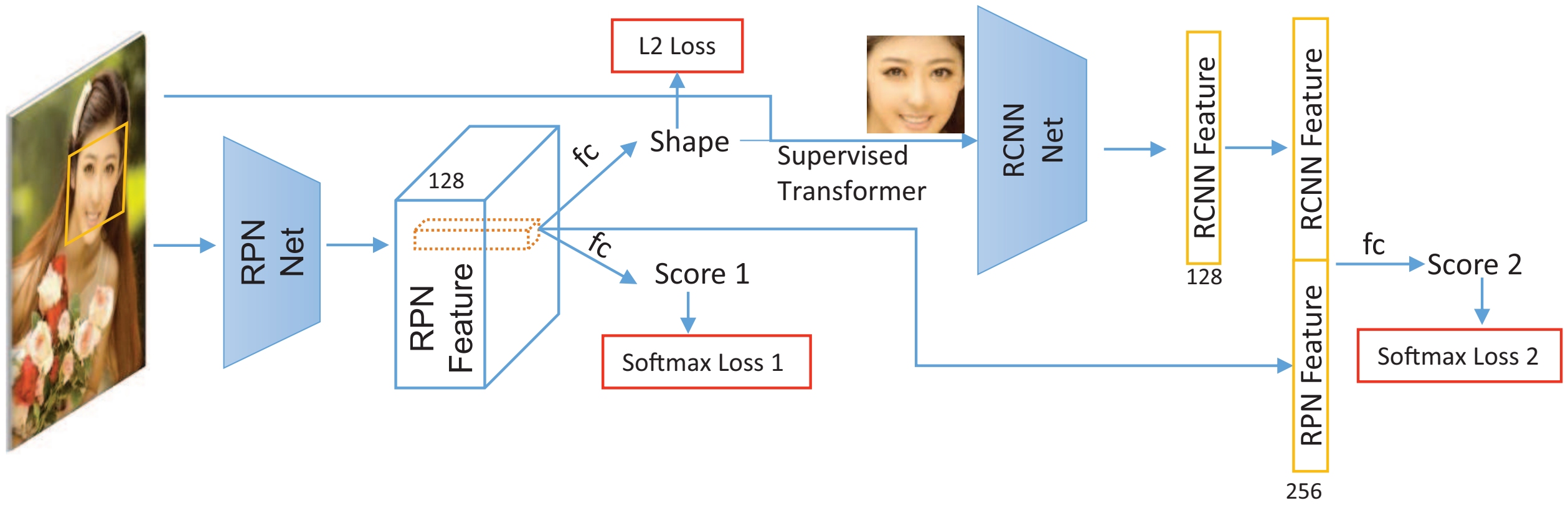

**Fig. 1.** Illustration of the structure of our Supervised Transformer Network.

Transformer Network conducts regression on the transformation parameters directly, and b) it is only supervised by the final recognition objective.

The proposed Supervised Transformer Network can efficiently run on the G-PU. However, in practice, the CPU is still the only choice in most situations. Therefore, we propose a region-of-interest (ROI) convolution scheme to make the run-time of the Supervised Transformer Network to be more efficient. It first uses a conventional boosting cascade to obtain a set of face candidate areas. Then, we combine these regions into irregular binary ROI mask. All DNN operations (including convolution, ReLU, pooling, and concatenation) are all processed inside the ROI mask, and hence significantly reduce the computation.

Our contributions are: 1) we proposed a new cascaded network named Supervised Transformer Network trained end-to-end for efficient face detection; 2) we introduced the supervised transformer layer, which enables to learn the optimal canonical pose to best differentiate face/non-face patterns; 3) we introduced a Non-top K suppression scheme, which can achieve better recall without sacrificing precision; 4) we introduced a ROI convolution scheme. It speeds up our detector 3x on CPU with little recall drop.

Our face detector outperformed the current best performing algorithms on several public benchmarks we evaluated, with real-time performance at 30 frames per second with VGA resolution.

## 2 Network Architecture

### 2.1 Overview

In this section, we will introduce the architecture of our proposed cascade network. As illustrated in Figure 1, the whole architecture consists of two stages. The first stage is a multi-task Region Proposal Network (RPN). It produces a set of candidate face regions along with associated facial landmarks. We conduct Non-top K suppression to only keep the candidate face regions with responses ranked in the top K in a local neighborhood.

The second stage starts with a Supervised Transformer layer, and then a RCNN to further verify if a face region is a true face or not. The transformer layer takes the facial landmarks and the candidate face regions, then warp the face regions into a canonical pose by mapping the detected facial landmarks into a set of canonical positions. This explicitly eliminates the effect of rotation and scale variation according to the facial points.

To make this clear, the geometric transformation are uniquely determined by the facial landmarks and the canonical positions. In our cascade network, both the prediction of the facial landmarks and the canonical positions are learned in the end-to-end training process. We call it a Supervised Transformer layer, as it receives supervision from two aspects. On one hand, the learning of the prediction model of the facial landmarks are supervised by the annotated ground-truth facial landmarks. On the other hand, the learning of both the canonical positions and the prediction model of the facial landmarks both are supervised by the final classification objective.

To make a final decision, we concatenate the fine-grained feature from the second-stage RCNN network and the global feature from the first-stage RPN network. The concatenated features are then put into a fully connected layer to make the final face/non-face arbitration. This concludes the whole architecture of our proposed cascade network.

### 2.2 Multi-task RPN

The design of the multi-task RPN is inspired by the JDA detector [16], which validated that face alignment is helpful to distinguish faces/non-faces. Our method is very straight forward. We use a RPN to simultaneous detect faces and associated facial landmarks. Our method is very similar to the work [20], except that our regression target is facial landmark locations, instead of bounding box parameters.

### 2.3 The supervised transformer layer

In this section, we describe the detail of the supervised transformer layer. As we know, similarity transformation was widely used in face detection and face recognition task to eliminate scale and rotation variation. The common practice is to train a prediction model to detect the facial landmarks, and then warp the face image to a canonical pose by mapping the facial landmarks to a set of manually specified canonical locations.

This process at least has two drawbacks: 1) one needs to manually set the canonical locations. Since the canonical locations determines the scale and offset of rectified face images, it often takes many try-and-errors to find a relatively good setting. This is not only time-consuming, but also suboptimal. 2) The learning of the prediction model for the facial landmark is supervised by the ground-truth facial landmark points. However, labeling ground-truth facial landmarks is a highly subjective process and hence prone to introducing noise.

We propose to learn both the canonical positions and the prediction of the facial landmarks end-to-end from the network with additional supervision information from the classification objective of the RCNN using end-to-end back propagation. Specifically, we use the following formula to define a similarity transformation, *i.e.*,

$$\begin{bmatrix} \bar{x}_i - m_{\bar{x}} \\ \bar{y}_i - m_{\bar{y}} \end{bmatrix} = \begin{bmatrix} a & b \\ -b & a \end{bmatrix} \begin{bmatrix} x_i - m_x \\ y_i - m_y \end{bmatrix}, \tag{1}$$

where $x_i, y_i$ are the detected facial landmarks, $\bar{x}_i, \bar{y}_i$ are the canonical positions, $m_*$ is the mean value of the corresponding variables, *e.g.*, $m_x = \frac{1}{N}\sum x_i$, $N$ is the number of facial landmarks, $a$ and $b$ are parameters of similarity transforms.

We found that this two parameters model is equivalent to the traditional four parameters, but much simpler in derivation and avoid problems of numerical calculation. After some straightforward mathematical derivation, we can obtain the least squares solution of the parameters, *i.e.*,

$$\begin{aligned} a &= \frac{c_1}{c_3} \\ b &= \frac{c_2}{c_3}. \end{aligned} \tag{2}$$

where

$$\begin{aligned} c_1 &= \sum \left((\bar{x}_i - m_{\bar{x}})(x_i - m_x) + (\bar{y}_i - m_{\bar{y}})(y_i - m_y)\right) \\ c_2 &= \sum \left((\bar{x}_i - m_{\bar{x}})(y_i - m_y) - (\bar{y}_i - m_{\bar{y}})(x_i - m_x)\right) \\ c_3 &= \sum \left((x_i - m_x)^2 + (y_i - m_y)^2\right). \end{aligned} \tag{3}$$

After obtaining the similarity transformation parameters, we can obtain the rectified image $\bar{I}$ given the original image $I$, using $\bar{I}(\bar{x}, \bar{y}) = I(x, y)$. Each point $(\bar{x}, \bar{y})$ in the rectified image can be mapped back to the original image space $(x, y)$ by

$$\begin{aligned} x &= \frac{a}{a^2 + b^2}(\bar{x} - m_{\bar{x}}) - \frac{b}{a^2 + b^2}(\bar{y} - m_{\bar{y}}) + m_x \\ y &= \frac{b}{a^2 + b^2}(\bar{x} - m_{\bar{x}}) + \frac{a}{a^2 + b^2}(\bar{y} - m_{\bar{y}}) + m_y. \end{aligned} \tag{4}$$

Since $x$ and $y$ may not be integers, bilinear interpolation is always used to obtain the value of $I(x, y)$. Therefore, we can calculate the derivative by the chain rule

$$\begin{aligned} \frac{\partial L}{\partial a} &= \sum_{\{\bar{x},\bar{y}\}} \frac{\partial L}{\partial \bar{I}(\bar{x}, \bar{y})} \frac{\partial \bar{I}(\bar{x}, \bar{y})}{\partial a} = \sum_{\{\bar{x},\bar{y}\}} \frac{\partial L}{\partial \bar{I}(\bar{x}, \bar{y})} \frac{\partial I(x, y)}{\partial a} \\ &= \sum_{\{\bar{x},\bar{y}\}} \frac{\partial L}{\partial \bar{I}(\bar{x}, \bar{y})} \left( \frac{\partial I(x, y)}{\partial x} \frac{\partial x}{\partial a} + \frac{\partial I(x, y)}{\partial y} \frac{\partial y}{\partial a} \right) \\ &= \sum_{\{\bar{x},\bar{y}\}} \frac{\partial L}{\partial \bar{I}(\bar{x}, \bar{y})} \left( I_x \frac{\partial x}{\partial a} + I_y \frac{\partial y}{\partial a} \right) \end{aligned} \tag{5}$$

where $L$ is the final classification loss and $\frac{\partial L}{\partial I(\bar{x},\bar{y})}$ is the gradient signals back propagated from the RCNN network. The $I_x$ and $I_y$ are horizontal and vertical gradient of the original image

$$
\begin{aligned}
I_x &= \beta_y(I(x_r, y_b) - I(x_l, y_b)) + (1 - \beta_y)(I(x_r, y_t) - I(x_l, y_t)) \\
I_y &= \beta_x(I(x_r, y_b) - I(x_r, y_t)) + (1 - \beta_x)(I(x_l, y_b) - I(x_l, y_t)).
\end{aligned} \tag{6}
$$

Here we use a bilinear interpolation, $\beta_x = x - \lfloor x \rfloor$ and $\beta_y = y - \lfloor y \rfloor$. $x_l = \lfloor x \rfloor, x_r = x_l + 1, y_t = \lfloor y \rfloor, y_b = y_t + 1$ are the left, right, top, bottom integer boundary of point $(x, y)$. Similarly, we can obtain the derivative of other parameters. Finally, we can obtain the gradient of the canonical positions of the facial landmarks, *i.e.*, $\frac{\partial L}{\partial \bar{x}_i}$ and $\frac{\partial L}{\partial \bar{y}_i}$. And the gradient with respect to the detected facial landmarks: $\frac{\partial L}{\partial x_i}$ and $\frac{\partial L}{\partial y_i}$. Please refer to the supplementary material for more detail.

The proposed Supervised Transformer layer is put between of the RPN and RCNN networks. In the end-to-end training, it automatically adjusts the canonical positions and guiding the detection of the facial landmarks such that the rectified image is more suitable for face/non-face classification. We will further illustrate this in the experiments.

### 2.4 Non-top K suppression

In RCNN [20, 17] based object detection, after the region proposals, non-maximum suppression (NMS) is always adopted to reduce the region candidate number for efficiency. However, the candidate with highest confidence score may be rejected by the later stage RCNN. Decreasing the NMS overlap threshold will bring in lots of useless candidates. This will make subsequent RCNN slow. Our idea is to keep K candidate regions with highest confidence for each potential face, since these samples are more promising for RCNN classifier. In the experiments part we will demonstrate that we can effectively improve the recall with the proposed Non-top K Suppression.

| **type** | **receptive field relationship** | **receptive field size** |
|---|---|---|
| conv1 $(7 \times 7, 2)$ | 2k+5 | 85 |
| max pool $(2 \times 2, 2)$ | 2k | 40 |
| conv 2a $(1 \times 1, 1)$ | k | 20 |
| conv 2b $(3 \times 3, 1)$ | k+2 | 20 |
| max pool $(2 \times 2, 2)$ | 2k | 18 |
| inception 3a | k+4 | 9 |
| inception 3b | k+4 | 5 |

**Table 1.** RPN network structure

### 2.5 Multi-granularity feature combination

Some works have revealed that joint features from different spatial resolutions or scales will improve accuracy [18]. The most straight-forward way may be combining several RCNN networks with different input scales. However, this approach will obviously increase the computation complexity significantly.

In our end-to-end network, the details of the RPN network structure is shown in Table 1. There are 3 convolution and 2 inception layers in our RPN network. Therefore, we can calculate that its receptive field size is 85. While the target face size is $36 \sim 72$ pixels. Therefore, our RPN takes advantage of the surrounding contextual information around face regions. On the other hand, the RCNN network focuses more on the rotation and scale variation fine grained detail in the inner face region. So we concatenate these two features in an end-to-end training architecture, which makes the two parts more complementary. Experiments demonstrate that this kind of joint feature can significantly improve the face detection accuracy. Besides, the proposed method is much more efficient.

## 3 The ROI convolution

### 3.1 Motivation

As a practical face detection algorithm, real-time performance is very important. However, the heavy computation incurred at test phase using DNN-based models often make them impractical in real-world systems. That is the reason why current DNN-based models heavily rely on a high-end GPU to increase the runtime performance. However, high-end GPU is not often available in commodity computing system, so most often, we still need to run the DNN model with a CPU. However, even using a high-end CPU with highly optimized code, it is still about 4 times slower than the runtime speed on a GPU [21]. More importantly, for portable devices, such as phones and tablets, mostly have low-end CPUs only, it is necessary to accelerate the test-phase performance of DNNs.

In a typical DNN, the convolutional layers are the most computationally expensive and often take up about more than 90% of the time in runtime. There were some works attempted to reduce the computational complexity of convolution layer. For example, Jaderberg *et al.* [22] applied a sparse decomposition to reconstruct the convolutional filters. Some other works [23, 24] assume that the convolutional filters are approximately low-rank along certain dimensions, and can be approximately decomposed into a series of smaller filters. Our detector may also benefit from these model compression techniques.

Nevertheless, we propose a more practical approach to accelerate the runtime speed of our proposed Supervised Transformer Network for face detection. Our main idea is to use a conventional cascade based face detector to quickly reject non-face regions and obtain a *binary ROI mask*. The ROI mask has the same size as the input. The background area is represented by 0 and the face area is represented by 1. The DNN convolution is only computed within the region marked as 1, ignoring all other regions. Because most regions did not participate

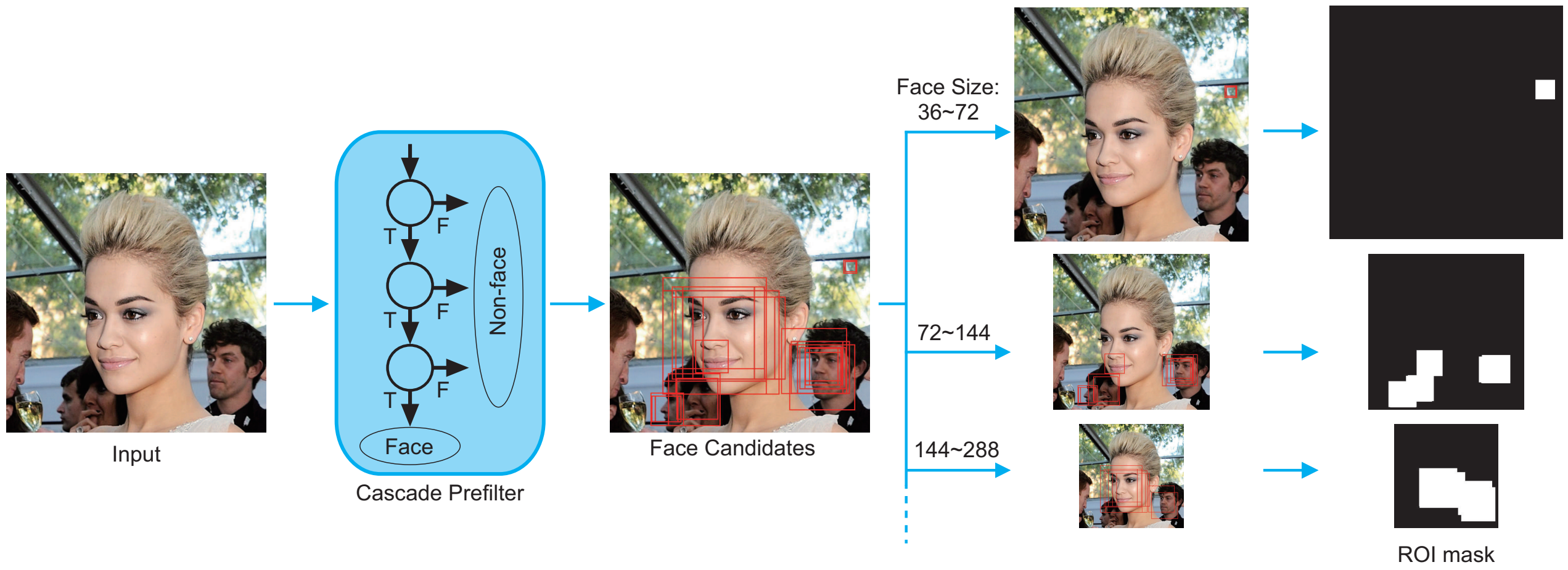

**Fig. 2.** Illustration of the ROI mask

in the calculation, we can greatly reduce the amount of computation in the convolution layers.

We want to emphasize that our method is different to those RCNN based algorithm [17, 25] which treated each candidate region independently. In those models, features in the overlap subregions will be calculated repeatedly. Instead, we use the ROI masks, so that different samples can share the feature in the overlapping area. It effectively reduces the computational cost by further avoiding repeated operations. Meanwhile, in the following section, we will introduce the implementation details of our ROI convolution. Similar to Caffe [26], we also take advantage of the matrix multiplication in the BLAS library to obtain almost a linear speedup.

### 3.2 Implementation details

**Cascade pre-filter**. As shown in Figure 2, we use a cascade detector as a pre-filter. It is basically a variant of the Volia-Jones's detector [1], but it has more weak classifiers and is trained with more data. Our boosted classifier is consisted of 1000 weak classifiers. Different form [1], we adopted a boosted fern [27] as the weaker classifier, since a fern is more powerful than using a single Haar feature based decision stump, and more efficient than boosted tree on CPUs. For completeness, we briefly describe our implementation.

Each fern contains 8 binary nodes. The splitting function is to compare the difference of two image pixel values in two different locations with a threshold, *i.e.*,

$$s_i = \begin{cases} 1 & p(x_{1_i}, y_{1_i}) - p(x_{2_i}, y_{2_i}) < \theta_i \\ 0 & otherwise \end{cases} \tag{7}$$

where $p$ is the image patch. The patch size is fixed to 32 in our experiments. The $(x_{1_i}, y_{1_i}, x_{2_i}, y_{2_i}, \theta_i)$ are fern parameters learned from training data. Each fern splits the data space into $2^8 = 256$ partitions. We use a Real-Boost algorithm for the cascade classification learning. In each space partition, the classification

score is computed as

$$\frac{1}{2}\log\left(\frac{\sum_{\{i\in piece\bigcap y_i=1\}} w_i}{\sum_{\{i\in piece\bigcap y_i=0\}} w_i}\right), \tag{8}$$

where the enumerator and denominator are the sum of the weights of positive and negative samples in the space partition, respectively.

**The ROI mask**. After we obtain some candidate face regions, we will group them according to their sizes. The maximum size is twice larger than the minimum size in each group. Since the smallest face size can be detected by the proposed DNN based face detector is $36 \times 36$ pixels, the first group contains the face size between 36 to 72 pixels. While the second ground contains the face size between 72 to 144, and so on (as shown in Figure 2).

It should be noted that, beginning from the second group, we need to down-sample the image, such that the candidate face size in the image is always maintained between 36 to 72 pixels. Besides, in order to retain some of the background information, we will double the side length of each candidate. But the side length will not exceed the receptive field size (85) of the following DNN face detector. Finally, we set the ROI mask according to the sizes and positions of the candidate boxes in each group.

We use this grouping strategy for two reasons. First, when there is a face almost filling the whole image, we do not have to deal with the full original image size. Instead, it will be down-sampled to a quite small resolution, so we can more effectively reduce the computation cost. Secondly, since the following DNN detector only need to handle twice the scale variation, this is induces a great advantage when compared with the RPN in [20], which needs to handle all scale changes. This advantage allows us to use a relatively cheaper network for the DNN-based detection.

Besides, such a sparse pyramid structure will only increase about 33% ($\frac{1}{2^2} + \frac{1}{4^2} + \frac{1}{8^2} \cdots \approx \frac{1}{3}$) computation cost when compared with the computational cost at the base scale.

**Details of the ROI convolution**. There are several ways to implement the convolutions efficiently. Currently, the most popular method is to transform the convolutions into a matrix multiplication. As described in [28] and implemented in Caffe [26], this can be done by firstly reshaping the filter tensor into a matrix $F$ with dimensions $CK^2 \times N$, where $C$ and $N$ are input and output channel numbers, and $K$ is the filter width/height.

We can subsequently gather a data matrix by duplicating the original input data into a matrix $D$ with dimensions $WH \times CK^2$, $W$ and $H$ are output width and height. The computation can then be performed with a single matrix multiplication to form an output matrix $O = DF$ with dimension $WH \times N$. This matrix multiplication can be efficiently calculated with optimized linear algebra libraries such as BLAS.

Our main idea in ROI convolution is to only calculate the area marked as 1 (*a.k.a*, the ROI regions), while skipping other regions. According to the ROI mask, we only duplicate the input patches whose centers are marked as 1. So the input data become a matrix $D'$ with dimensions $M \times CK^2$, where $M$ is the number of non-zero entries in the ROI mask. Similarly, we can then use matrix multiplication to obtain the output $O' = D'F$ with dimension $M \times CK^2$. Finally, we put each row of $O'$ to the corresponding channel of the output. The computation complexity of ROI convolution is $MCK^2N$. Therefore, we can linearly decrease the computation cost according to the mask sparsity.

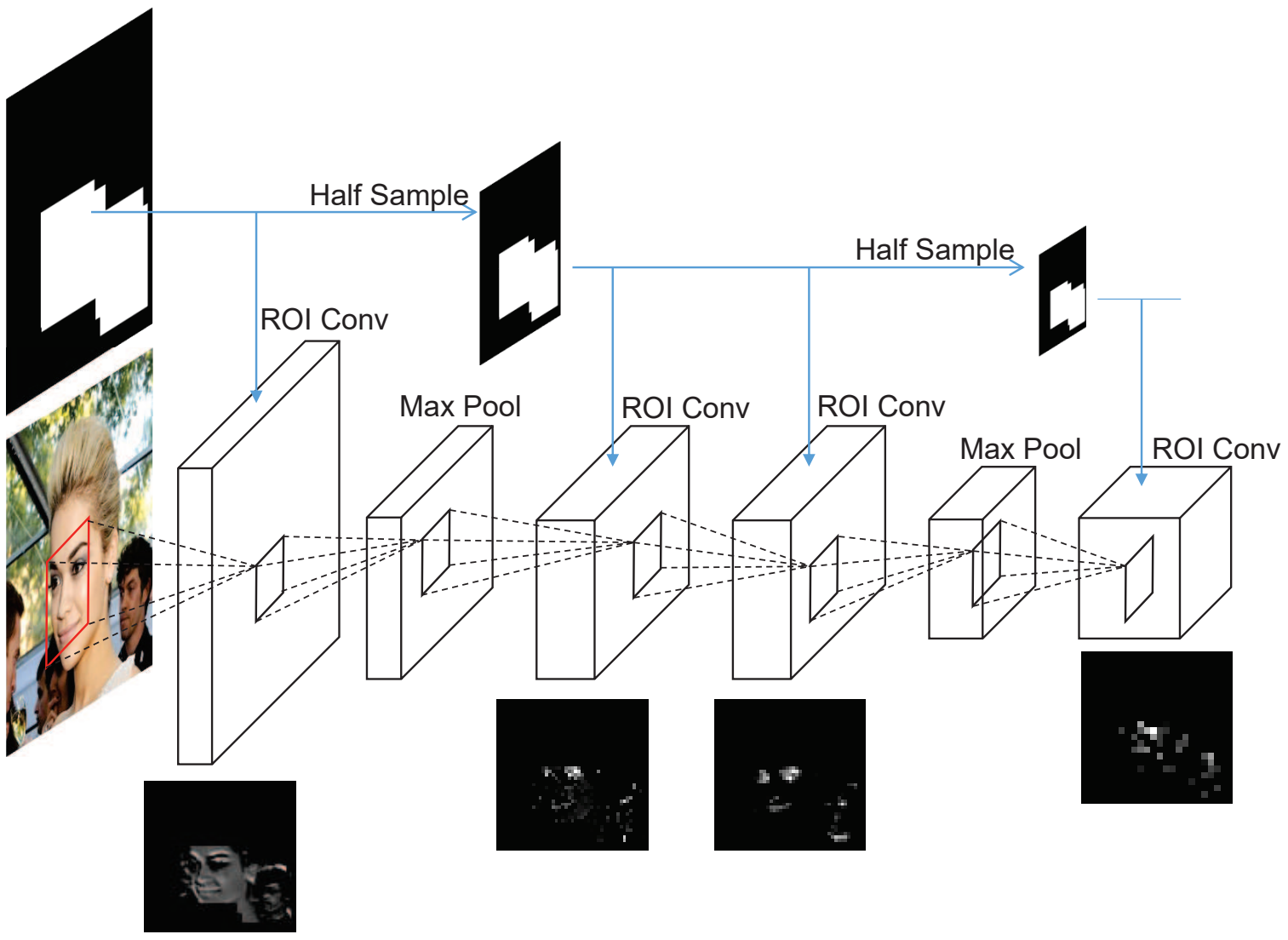

**Fig. 3.** Illustration of the ROI convolution.

As illustrated in Figure 3, we only apply the ROI convolution in the test phase. We replace all convolution layers into ROI convolution layers. After a max pooling, the size of the input will be halved. So we also half sample the ROI mask, such that their size can be matched. The original DNN detector can run at 50 FPS on GPU and 10 FPS on CPU for a VGA image. With ROI convolution, it can speed up to 30 FPS on CPU with little accuracy loss.

## 4 Experiments

In this section, we will experimentally validate the proposed method. We collected about $400K$ face images from the web with various variations as positive training samples. These images are exclusive from FDDB [29], AFW [8] and PASCAL [30] datasets. We labeled all faces with 5 facial points (two eyes center, nose tip, and two mouth corners). For the negative training samples, we use the Coco database [31]. This dataset has pixel level annotations of various objects, including people. Therefore, we covered all person areas with random color blocks, and ensure that no samples are drawn from those colored regions in these images. We use more than $120K$ images (including 2014 training and validation data) for the training. Some sample images are shown in Fig. 4.

We use GoogleNet in both the RPN and RCNN networks. The network structure is similar to that in FaceNet [32], but we cut all the convolution kernel number in half for efficiency. Moreover, we only include two inception layers in RPN network (as shown in Table 1) and the input size of RCNN network is 64.

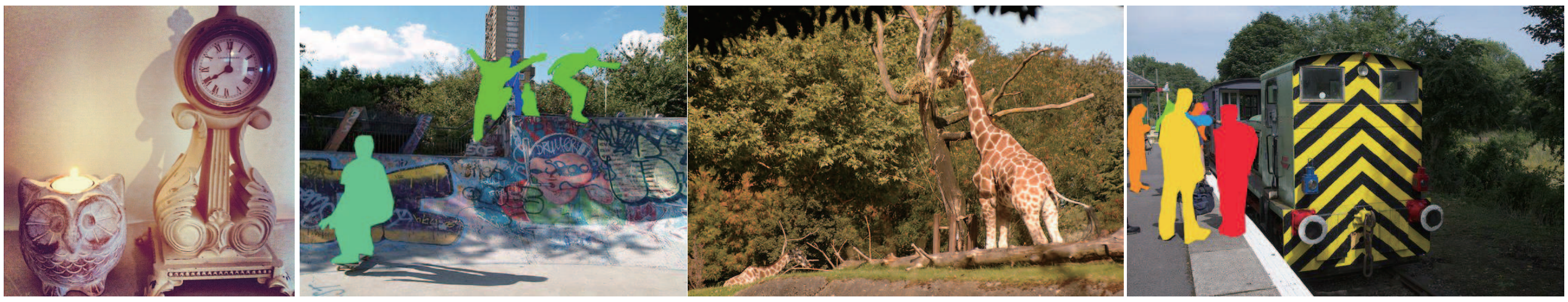

**Fig. 4.** Illustration of our negative training sample. We covered all person area with random color blocks in Coco [31] dataset and ensured that no positive training samples are drawn from these regions in these images.

In order to avoid the initialization problem and improve the convergence speed, we first train the RPN network from random without the RCNN network. After the predicted facial landmarks are largely correct, we add the RCNN network and perform end-to-end training together. For evaluation, we use three challenging public datasets, *i.e.*, FDDB [29], AFW [8] and PASCAL faces [30]. All these three datasets are widely used as face detection benchmark. We employ the Intersection over Union (IoU) as the evaluation metric and fix the IoU threshold to 0.5.

### 4.1 Learning canonical position

In this part, we verify the effect of the Supervised Transformation in finding the best canonical position. We intentionally initialize the Supervised Transformation with three inappropriate canonical positions according to three settings, respectively, *i.e.*, too large, too small, or with offset. Then we perform the end-to-end training and record the canonical points position after 10K, 100K, 500K iterations.

As shown in Fig. 5, each row shows the canonical positions movement for one kind of initializations. We also place the image warp result besides its corresponding canonical points. We can observe that, for these three different kinds of initializations, they all eventually converge to a very close position setting after 500K iterations. It demonstrated that the proposed Supervised Transformer module is robust to the initialization. It automatically adjusts the canonical positions such that the rectified image is more suitable for face/non-face classification.

### 4.2 Ablative evaluation of various network components

As discussed in Sec. 2, our end-to-end cascade network is consisted of four notable parts, *i.e.*, the multi-task RPN, the Supervised Transformer, the multi-granularity feature combination, and non-top K suppression. The former three will affect the network structure of training, while the last one only appear in the test phase.

In order to separately study the effect of each part, we conduct an ablative study by removing one or more parts from our network structure and evaluate the new network with the same training and testing data. When removing the multi-task RPN, it means that we directly regress the face rectangle similar to [20], instead of facial points. Without the Supervised Transformer layer, we simply replace it with a standard similarity transformation without training with back propagation. Without the feature combination component means that we directly use the output of the RCNN features to make the finial decision. In the case that we removed multi-task RPN, there will be no facial points for Supervised Transformation or conventional similarity transformation. In this situation, we directly resize the face patch into $64 \times 64$ and fed it into a RCNN network.

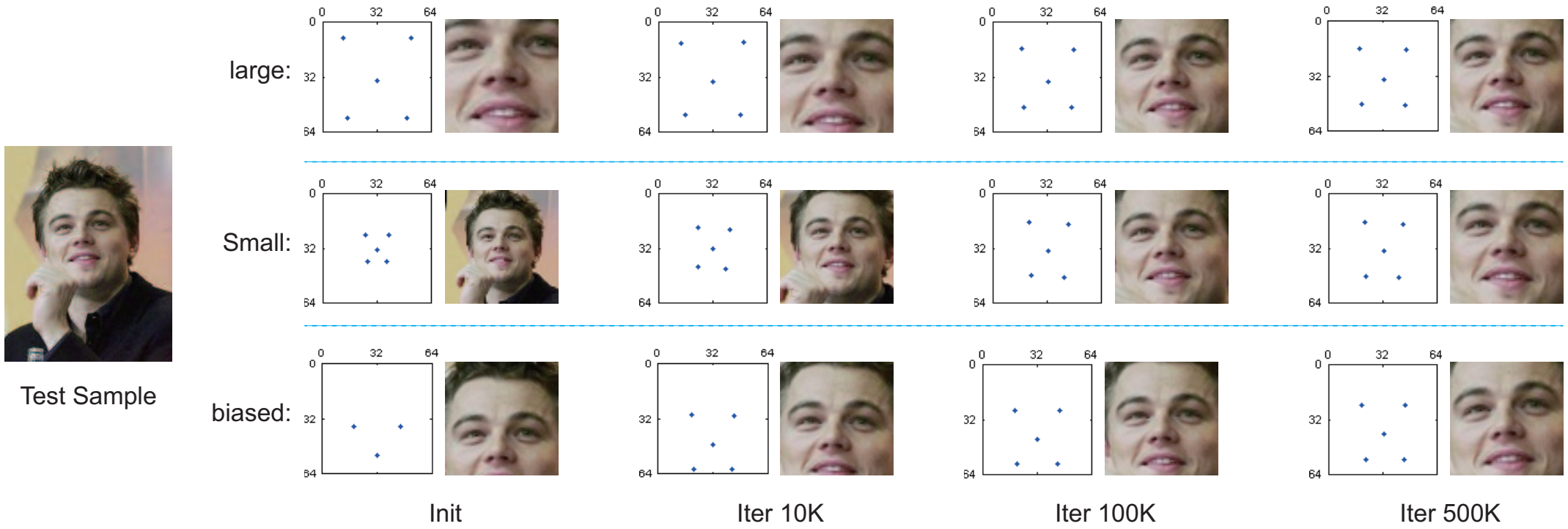

**Fig. 5.** Results of learning canonical positions.

| | | | | | | |
|---|---|---|---|---|---|---|
| Multi-task RPN | N | N | Y | Y | Y | Y |
| Supervised Transformer | / | / | N | Y | N | Y |
| Feature combination | N | Y | N | N | Y | Y |
| Recall Rate | 85.6% | 88.0% | 87.1% | 88.3% | 88.8% | 89.6% |

**Table 2.** Evaluation of the effect of three parts in training architecture.

There are 6 different ablative settings in total. We perform end-to-end training with the same training samples for all settings, and evaluate the recall rate on the FDDB dataset when the *false alarm* number is 10. We manually review the face detection results and add 67 unlabeled faces in the FDDB dataset to make sure all the false alarms are true. As shown in Table 2, multi-task RPN, Supervised Transformer, and feature combination will bring about 1%, 1%, and 2% recall improvement respectively. Besides, these three parts are complementary, remove any one part will cause a recall drop.

In the training phase, in order to increase the variation of training samples, we randomly select K positive/negative samples from each image for the RCNN network. However, in the test phase, we need to balance the recall rate with efficiency. Next, we will compare the proposed non-top K suppression with NMS in the testing phase,

We present a sample visual result of RPN, NMS and non-top K suppression in Fig. 6. We keep the same number of candidates for both NMS and Non-top K suppression ($K = 3$ in the visual result). We found that NMS tend to include too much noisy low confidence candidates. We also compare the PR curves of using all candidates, NMS, and non-top K suppression. Our non-top K suppression is very close to using all candidates, and achieved consistently better results than NMS under the same number of candidates.

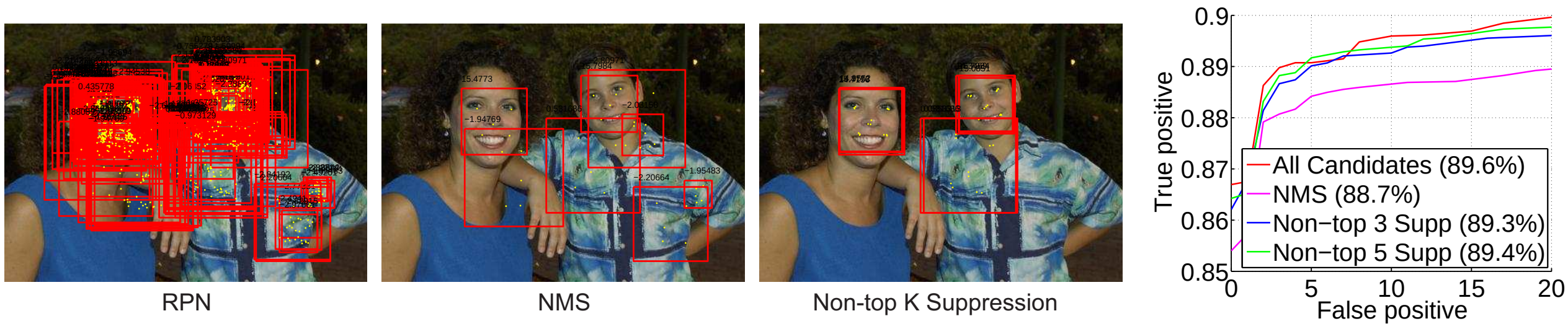

**Fig. 6.** Comparison of NMS and Non-Top K Suppression

| Pre-filter Threshold | $N/A$ | 0 | 1 | 2 | 3 |
|---|---|---|---|---|---|
| ROI Mask Sparsity | $N/A$ | 31.3% | 27.1% | 10.6% | 5.7% |
| Pre-filter Time (ms) | 0 | 12.1 | 12.0 | 12.0 | 11.9 |
| RPN Time (ms) | 98.2 (100%) | 33.9 (34.5%) | 24.2 (24.6%) | 11.0 (11.2%) | 8.1 (8.2%) |
| RCNN Time (ms) | 9.0 | 9.3 | 8.7 | 9.1 | 9.3 |
| Total Time (ms) | 107.2 | 55.3 | 44.8 | 32.1 | 29.3 |
| Recall Rate | 89.3% | 89.2% | 89.0% | 88.7% | 88.1% |

**Table 3.** Various results demonstrating the effects of ROI convolution.

### 4.3 The effect of ROI convolution

In this section, we will validate the acceleration performance of the proposed ROI convolution algorithm. We train the Cascade pre-filter with the same training data. By adjusting the classification threshold of the Cascade re-filter, we can obtain the ROI masks in different areas. Therefore, we can strike for the right balance between speed and accuracy.

We conduct the experiments on the FDDB database. We resized all images to 1.5 times of the original size, the resulting average photos resolution is approximately $640 \times 480$. We evaluate the ROI mask sparsity, run-time speed [1] of each part, and the recall rate when the false alarm number is 10 under different pre-filter threshold. We also compare with the standard network without ROI convolution. Non-top K ($K = 3$) suppression is adopted in all settings to make RCNN network more efficiency.

Table 3 shows the average ROI mask sparsity, testing speed of each part, and recall rate of each setting. Comparing the second row with the fourth row, it proves that we can linearly decrease the computation cost according to the mask sparsity. The last two rows show the recall rate and average test time of different settings. The original DNN detector can run at 10 FPS on CPU for a VGA image. With ROI convolution, it can speed up to 30 FPS on CPU. We can achieve about 3 times speed up with only 0.6% recall rate drop.

[1] All experiments use a single thread on an Intel i7-4770K CPU

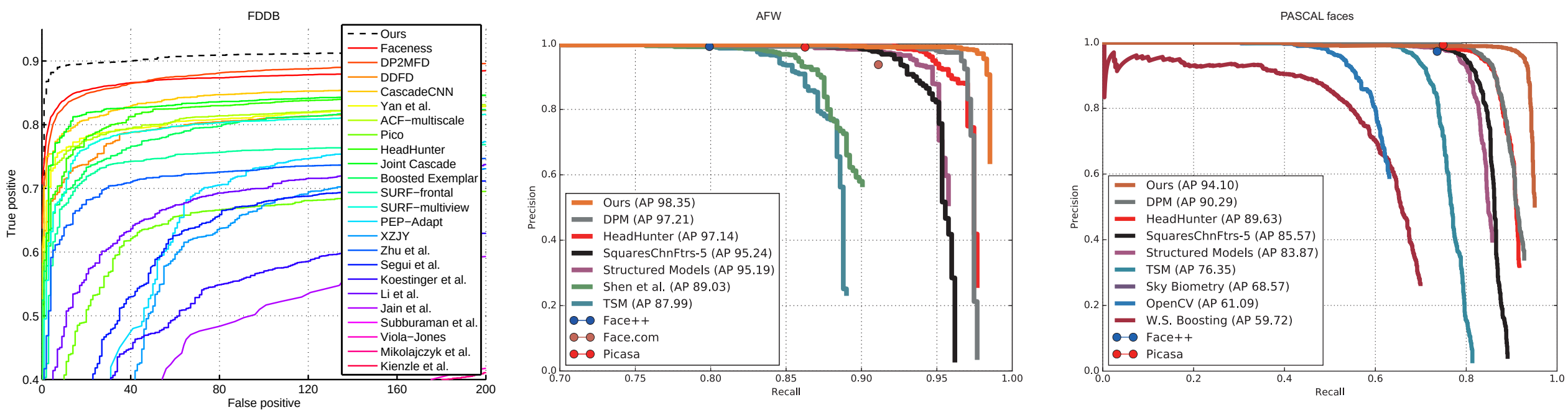

**Fig. 7.** Comparison with state-of-the-arts on the FDDB [29], AFW [8] and PASCAL faces [30] datasets.

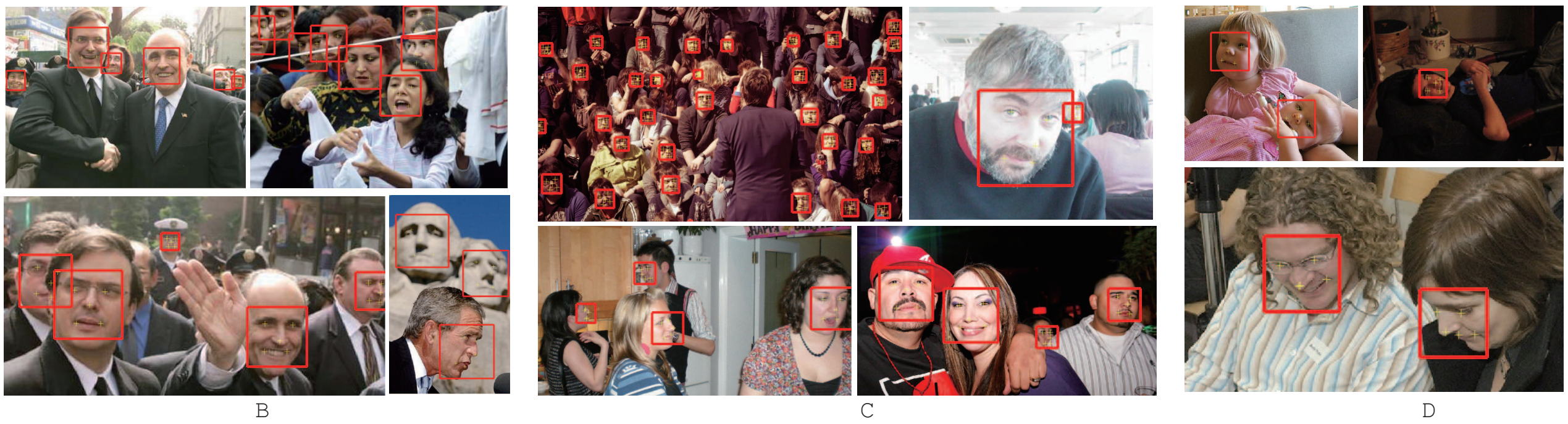

**Fig. 8.** Qualitative face detection results on (a) FDDB [29], (b) AFW [8], (c) PASCAL faces [30] datasets.

### 4.4 Comparing with state-of-the-art

We conduct face detection experiments on three benchmark datasets. On the FDDB dataset, we compare with all public methods [33, 8, 34, 35, 9, 36–40, 35, 10, 41, 42]. We regress the annotation ellipses with 5 facial points and ignore 67 unlabeled faces to make sure all false alarms are true. On the AFW and PASCAL faces datasets, we compare with (1) deformable part based methods, *e.g.* structure model [30] and Tree Parts Model (TSM) [8]; (2) cascade-based methods, *e.g.* Headhunter [4]; (3) commercial system, *e.g.* face.com, Face++ and Picasa. We learn a global regression from 5 facial points to face rectangles to match the annotation for each dataset, and use toolbox from [4] for the evaluation. Fig. 8 shows that our method outperforms all previous methods by a considerable margin.

## 5 Conclusion and future work

In this paper, we proposed a new Supervised Transformer Network for face detection. The superior performance on three challenge datasets shows its ability to learn the optimal canonical positions to best distinguish face/non-face patterns. We also introduced a ROI convolution, which speeds up our detector 3x on CPU with little recall drop. Our future work will explore how to enhance the ROI convolution so that it does not incur additional drops in recall.

## References

1. Viola, P., Jones, M.: Rapid object detection using a boosted cascade of simple features. In: Computer Vision and Pattern Recognition, 2001. CVPR 2001. Proceedings of the 2001 IEEE Computer Society Conference on. Volume 1., IEEE (2001) 511–518
2. Li, S.Z., Zhu, L., Zhang, Z., Blake, A., Zhang, H., Shum, H.: Statistical learning of multi-view face detection. In: European Conference on Computer Vision. (2002) 67–81
3. Wu, B., Ai, H., Huang, C., Lao, S.: Fast rotation invariant multi-view face detection based on real adaboost. In: Automatic Face and Gesture Recognition. (2004) 79–84
4. Mathias, M., Benenson, R., Pedersoli, M., Van Gool, L.: Face detection without bells and whistles. In: European Conference on Computer Vision. (2014) 720–735
5. Viola, M., Jones, M.J., Viola, P.: Fast multi-view face detection. In: TR2003-96. (2003)
6. Huang, C., Ai, H., Li, Y., Lao, S.: Vector boosting for rotation invariant multi-view face detection. In: Computer Vision, 2005. ICCV 2005. Tenth IEEE International Conference on. Volume 1. (Oct 2005) 446–453 Vol. 1
7. Huang, C., Ai, H., Li, Y., Lao, S.: High-performance rotation invariant multiview face detection. Pattern Analysis and Machine Intelligence, IEEE Transactions on **29**(4) (April 2007) 671–686
8. Zhu, X., Ramanan, D.: Face detection, pose estimation, and landmark localization in the wild. In: Computer Vision and Pattern Recognition (CVPR), 2012 IEEE Conference on, IEEE (2012) 2879–2886
9. Li, H., Hua, G., Lin, Z., Brandt, J., Yang, J.: Probabilistic elastic part model for unsupervised face detector adaptation. In: The IEEE International Conference on Computer Vision (ICCV). (2013)
10. Yang, S., Luo, P., Loy, C.C., Tang, X.: From facial parts responses to face detection: A deep learning approach. In: Proceedings of the IEEE International Conference on Computer Vision. (2015) 3676–3684
11. Dollar, P., Appel, R., Belongie, S., Perona, P.: Fast feature pyramids for object detection. Pattern Analysis and Machine Intelligence, IEEE Transactions on **36**(8) (Aug 2014) 1532–1545
12. Yang, B., Yan, J., Lei, Z., Li, S.Z.: Convolutional channel features for pedestrian, face and edge detection. CoRR **abs/1504.07339** (2015)
13. Shen, X., Lin, Z., Brandt, J., Wu, Y.: Detecting and aligning faces by image retrieval. In: Computer Vision and Pattern Recognition (CVPR), 2013 IEEE Conference on. (June 2013) 3460–3467
14. Farfade, S.S., Saberian, M.J., Li, L.J.: Multi-view face detection using deep convolutional neural networks. In: Proceedings of the 5th ACM on International Conference on Multimedia Retrieval. ICMR '15, New York, NY, USA, ACM (2015) 643–650
15. Li, H., Lin, Z., Shen, X., Brandt, J., Hua, G.: A convolutional neural network cascade for face detection. In: Computer Vision and Pattern Recognition (CVPR), 2015 IEEE Conference on. (June 2015) 5325–5334
16. Chen, D., Ren, S., Wei, Y., Cao, X., Sun, J.: Joint cascade face detection and alignment. In: Proceedings of the European Conference on Computer Vision (ECCV). (2014)
17. Girshick, R., Donahue, J., Darrell, T., Malik, J.: Rich feature hierarchies for accurate object detection and semantic segmentation. In: Computer Vision and Pattern Recognition (CVPR), 2014 IEEE Conference on. (June 2014) 580–587

18. Bell, S., Zitnick, C.L., Bala, K., Girshick, R.: Inside-outside net: Detecting objects in context with skip pooling and recurrent neural networks. arXiv preprint arXiv:1512.04143 (2015)
19. Jaderberg, M., Simonyan, K., Zisserman, A., et al.: Spatial transformer networks. In: Advances in Neural Information Processing Systems. (2015) 2008–2016
20. Ren, S., He, K., Girshick, R., Sun, J.: Faster r-cnn: Towards real-time object detection with region proposal networks. In: Advances in Neural Information Processing Systems. (2015) 91–99
21. Vanhoucke, V., Senior, A., Mao, M.Z.: Improving the speed of neural networks on cpus. In: Deep Learning and Unsupervised Feature Learning Workshop, NIPS 2011. (2011)
22. Liu, B., Wang, M., Foroosh, H., Tappen, M., Penksy, M.: Sparse convolutional neural networks. In: Computer Vision and Pattern Recognition (CVPR), 2015 IEEE Conference on. (June 2015) 806–814
23. Jaderberg, M., Vedaldi, A., Zisserman, A.: Speeding up convolutional neural networks with low rank expansions. In: Proceedings of the British Machine Vision Conference, BMVA Press (2014)
24. Zhang, X., Zou, J., Ming, X., He, K., Sun, J.: Efficient and accurate approximations of nonlinear convolutional networks. In: Computer Vision and Pattern Recognition (CVPR), 2015 IEEE Conference on. (June 2015) 1984–1992
25. Zhang, C., Zhang, Z.: Improving multiview face detection with multi-task deep convolutional neural networks. In: Applications of Computer Vision (WACV), 2014 IEEE Winter Conference on. (March 2014) 1036–1041
26. Jia, Y., Shelhamer, E., Donahue, J., Karayev, S., Long, J., Girshick, R., Guadarrama, S., Darrell, T.: Caffe: Convolutional architecture for fast feature embedding. In: Proceedings of the ACM International Conference on Multimedia, ACM (2014) 675–678
27. Ozuysal, M., Fua, P., Lepetit, V.: Fast keypoint recognition in ten lines of code. In: Computer Vision and Pattern Recognition, 2007. CVPR'07. IEEE Conference on, Ieee (2007) 1–8
28. Chellapilla, K., Puri, S., Simard, P.: High performance convolutional neural networks for document processing. In: Tenth International Workshop on Frontiers in Handwriting Recognition, Suvisoft (2006)
29. Jain, V., Learned-Miller, E.: Fddb: A benchmark for face detection in unconstrained settings. Technical Report UM-CS-2010-009, University of Massachusetts, Amherst (2010)
30. Yan, J., Zhang, X., Lei, Z., Li, S.Z.: Face detection by structural models. Image and Vision Computing **32**(10) (2014) 790–799
31. Lin, T.Y., Maire, M., Belongie, S., Hays, J., Perona, P., Ramanan, D., Dollár, P., Zitnick, C.L.: Microsoft coco: Common objects in context. In: Computer Vision–ECCV 2014. Springer (2014) 740–755
32. Schroff, F., Kalenichenko, D., Philbin, J.: Facenet: A unified embedding for face recognition and clustering. In: Proceedings of the IEEE Conference on Computer Vision and Pattern Recognition. (2015) 815–823
33. Wu, B., Ai, H., Huang, C., Lao, S.: Fast rotation invariant multi-view face detection based on real adaboost. In: Automatic Face and Gesture Recognition, 2004. Proceedings. Sixth IEEE International Conference on. (May 2004) 79–84
34. Shen, X., Lin, Z., Brandt, J., Wu, Y.: Detecting and aligning faces by image retrieval. In: Computer Vision and Pattern Recognition (CVPR), 2013 IEEE Conference on. (June 2013) 3460–3467

35. Li, H., Lin, Z., Brandt, J., Shen, X., Hua, G.: Efficient boosted exemplar-based face detection. In: Computer Vision and Pattern Recognition (CVPR), 2014 IEEE Conference on. (June 2014) 1843–1850
36. Li, J., Zhang, Y.: Learning surf cascade for fast and accurate object detection. In: Computer Vision and Pattern Recognition (CVPR), 2013 IEEE Conference on. (June 2013) 3468–3475
37. Jain, V., Learned-Miller, E.: Online domain adaptation of a pre-trained cascade of classifiers. In: Computer Vision and Pattern Recognition (CVPR), 2011 IEEE Conference on, IEEE (2011) 577–584
38. Subburaman, V.B., Marcel, S.: Fast bounding box estimation based face detection. In: ECCV, Workshop on Face Detection: Where we are, and what next? (2010)
39. Mikolajczyk, K., Schmid, C., Zisserman, A.: Human detection based on a probabilistic assembly of robust part detectors. In: Computer Vision-ECCV 2004. Springer (2004) 69–82
40. Yan, J., Lei, Z., Wen, L., Li, S.: The fastest deformable part model for object detection. In: Computer Vision and Pattern Recognition (CVPR), 2014 IEEE Conference on. (June 2014) 2497–2504
41. Ranjan, R., Patel, V.M., Chellappa, R.: A deep pyramid deformable part model for face detection. In: Biometrics Theory, Applications and Systems (BTAS), 2015 IEEE 7th International Conference on, IEEE (2015) 1–8
42. Farfade, S.S., Saberian, M.J., Li, L.J.: Multi-view face detection using deep convolutional neural networks. In: Proceedings of the 5th ACM on International Conference on Multimedia Retrieval, ACM (2015) 643–650